\documentclass[10pt,conference]{IEEEtran}
\IEEEoverridecommandlockouts
\usepackage{cite}
\usepackage{amsmath,amssymb,amsfonts}
\usepackage{algorithmic}
\usepackage{graphicx}
\usepackage{textcomp}
\usepackage{xcolor}
\usepackage{booktabs}
\usepackage{mathtools}
\usepackage[english]{babel}
\usepackage{multirow}
\usepackage{subfig}
\usepackage{url}
\usepackage{hyperref}
\usepackage{mathabx}
\usepackage{amsmath}
\usepackage{amssymb}
\usepackage{stmaryrd}
\usepackage[export]{adjustbox}
\usepackage[english,ruled,vlined,linesnumbered]{algorithm2e}
\usepackage[justification=centering]{caption}
\newtheorem{proof}{Proof}
\newtheorem{property}{Property}

\newcommand{\SplitBasedSelection}{\mbox{\texttt{SplitSD4X}}}
\newcommand{\generateNeighbors}{\mbox{GenerateNeighbors}}
\newcommand{\comment}[1]{}

\def\BibTeX{{\rm B\kern-.05em{\sc i\kern-.025em b}\kern-.08em
    T\kern-.1667em\lower.7ex\hbox{E}\kern-.125emX}}

\AtBeginDocument{
  \catcode`_=12
  \begingroup\lccode`~=`_
  \lowercase{\endgroup\let~}\sb
  \mathcode`_="8000
}

\newtheorem{problem}{Problem}

\begin{document}

\title{Interpretable Summaries of Black Box Incident Triaging with Subgroup Discovery} 

\author{
Youcef Remil\textsuperscript{1,2,4},
Anes Bendimerad\textsuperscript{2},
Marc Plantevit\textsuperscript{3},
C{\'{e}}line Robardet\textsuperscript{1},
Mehdi Kaytoue\textsuperscript{1,2}

\\\\
\textsuperscript{1}  Univ Lyon, INSA Lyon, CNRS, LIRIS UMR 5205, F-69621, Lyon, France
\\
\textsuperscript{2}  Infologic, 99 avenue de Lyon, F-26500, Bourg-L{\`{e}}s-Valence, France
\\
\textsuperscript{3}  Univ Lyon, CNRS, LIRIS UMR 5205, F-69622, Lyon, France
\\
\textsuperscript{4} \texttt{yre@infologic.fr}}

\maketitle

\begin{abstract}
The need of predictive maintenance comes with an increasing number of incidents reported by monitoring systems and equipment/software users. In the front line, on-call engineers (OCEs) have to quickly assess the degree of severity of an incident and decide which service to contact for corrective actions. To automate these decisions, several predictive models have been proposed, but the most efficient models are opaque (say, black box), strongly limiting their adoption. In this paper, we propose an efficient black box model based on 170K incidents reported to our company over the last 7 years and emphasize on the need of automating triage when incidents are massively reported on thousands of servers running our product, an ERP. Recent developments in eXplainable Artificial Intelligence (XAI) help in providing global explanations to the model, but also, and most importantly, with local explanations for each model prediction/outcome. Sadly, providing a human with an explanation for each outcome is not conceivable when dealing with an important number of daily predictions. To address this problem, we propose an original data-mining method rooted in Subgroup Discovery, a pattern mining technique with the natural ability to group objects that share similar explanations of their black box predictions and provide a description for each group. We evaluate this approach and present our preliminary results which give us good hope towards an effective OCE's adoption. We believe that this approach provides a new way to address the problem of model agnostic outcome explanation.

\end{abstract}

\section{Introduction}\label{sec:introduction}

\begin{figure*}
\centering
\includegraphics[width=0.95\textwidth]{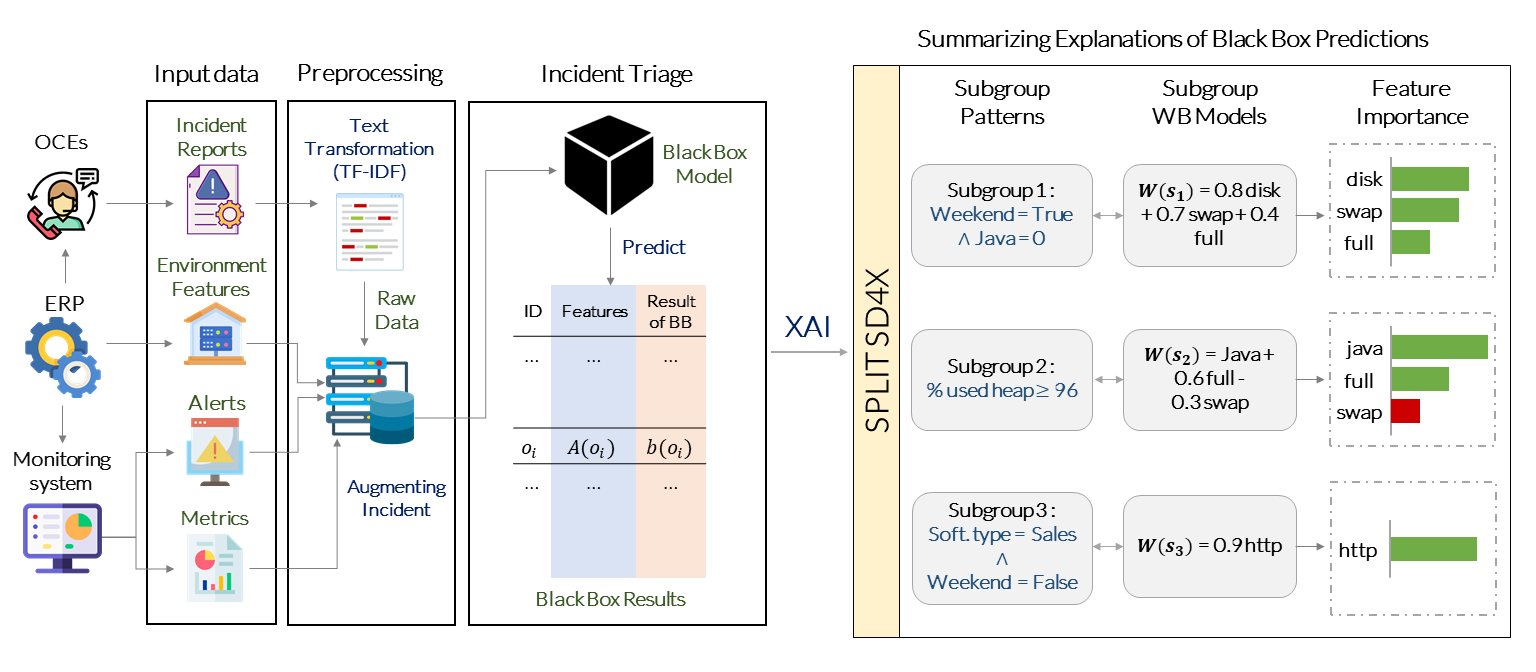}
\caption{\label{fig:generalSchema} Overview of our explainable incident triage framework: (1)~Input data are processed to extract relevant features, (2)~a black box model uses these features to provide an accurate incident triage, (3)~contextualized groups of black box predictions are explained using a Subgroup Discovery approach.}
\end{figure*}

Many industries are still moving towards digitization: software such as Enterprise Resource Planning (ERP) are at the center of this revolution. ERPs are directly connected to (distant) factories and their equipments. Software layers allow the user to interact with equipments and data for achieving their daily tasks in the most efficient way, where AI increasingly comes at play. It follows that the maintenance of an ERP becomes more and more complex as involving always more interrelated physical, software and business components. Collecting (big) data with supervision systems, integrated in the ERP for contextualizing them as much as possible (sometimes reflected as \textit{observability}), is key for the early detection of incidents. Incidents are actually of various natures: (i) a predicted incident  based on models built upon historical data with a certain probability, (ii) an actual incident not yet reported by the end-user but by the supervision system with a certain degree of severity and impact on the production and end-user, and (iii) of course, an incident directly reported by the end-user, but generally poorly described and contextualized. Some incidents are insignificant  \textit{per se} but severe when co-occurring with others. 

It follows an increasing number of incidents reported by monitoring systems and equipment/software users (tickets, phone calls, etc). In the front line, on-call engineers (OCEs) have to quickly assess the degree of severity of an incident and decide which service to contact for a palliative and/or curative actions. To speed up and improve these decisions, several predictive models have been proposed for \textit{incident triage} in the last few years, aiming at predicting the service to target.
\cite{DBLP:conf/kbse/0003HL0HGXDZ19} have addressed this problem with a Deep Learning approach and demonstrated its superiority over existing methods proposed for the similar problem of bug triage~\cite{DBLP:conf/sigsoft/LeeHLKJ17}. Few other techniques have been also designed to automate incident triage~\cite{DBLP:conf/kdd/PhamJDOJ20,DBLP:conf/sigcomm/GaoYMFLBAWLMYA20,wang2021fast} and are mainly based on DNN and ensemble models. Unfortunately, such methods build opaque models, qualified as \textit{black box}, as they do not convey any explanation of their output (outcome or prediction) to the end user. An important challenge to successfully automate incident triage is to gain practitioners trust, providing them with an explanation on each model outcome, which can be also part of the incident resolution, as it helps with root cause analysis. 

The popularity of black box prediction models combined with the crucial need of transparency in many decision processes has conducted to an unprecedented interest for \textit{eXplainable Artificial Intelligence} (XAI). Some methods aim at \textit{globally} explaining the internal logic of the black box, while others provide \textit{local} explanations for specific outcomes. Global explanations are generally provided by learning a potentially interpretable model (e.g., decision tree, classification rules) that mimics the predictions of the original black box model~\cite{DBLP:conf/aistats/HaraH18,DBLP:journals/pr/KrishnanSB99,DBLP:journals/npl/AugastaK12,DBLP:journals/titb/HanLYPC15}. However, it is often not feasible to explain all the logic as the model can be extremely complex, and global explanations may easily miss some predictions given to the end user. Thus, most of recent work focus on local methods that explain each outcome of a black box model independently. These methods provide a local explanation for the prediction of a specific object that can only be interpreted locally. 
Some of these approaches are designed for a specific model such as DNN~\cite{DBLP:journals/corr/SimonyanVZ13,DBLP:conf/icml/ShrikumarGK17,DBLP:conf/icml/SundararajanTY17,DBLP:conf/eccv/ZeilerF14} and GNN~\cite{DBLP:conf/nips/YingBYZL19}. On the other hand, many other methods~\cite{DBLP:conf/kdd/Ribeiro0G16,DBLP:conf/pakdd/GuidottiMC19,DBLP:journals/corr/abs-2002-03746} are able to explain the outcome of any model (agnostic). They are often based on the generation of a local neighborhood around the object to explain, and on the training of an interpretable model that mimic well the black box on this neighborhood. The drawback of outcome explanation methods is the large number of provided explanations if applied to a large number of outcomes. Analyzing individual explanations becomes very time consuming for the end-user. Few methods have been proposed to address this issue, by grouping similar explanations into clusters~\cite{DBLP:conf/aies/IbrahimLMP19} or by selecting a subset of representative explanations~\cite{DBLP:conf/kdd/Ribeiro0G16}. These two methods can thus give a picture of the different possible explanations of a model, but they do not provide to the user the contexts in which each explanation holds.

In this article, we first present the maintenance platform of our ERP allowing us to motivate the need of automatically triaging incidents of different natures. Then, we propose an efficient black box model based on 170K user reported incidents that we dealt with over the last 7 years. This motivates our third contribution which consists in an original approach that summarizes local explanations of black box predictions. This method is rooted in Subgroup Discovery~\cite{DBLP:books/mit/fayyadPSU96/Klosgen96,DBLP:journals/widm/Atzmueller15} to group predicted objects into subgroups that support the same explanation. Instead of providing a specific explanation for the prediction of each object, (i) we group objects into a controlled number of subgroups, and for each subgroup (ii) we provide an explanation that holds for all of its objects. Each of these subgroups is also associated with a description (a restriction on attribute values) that separates it exactly from the rest of the dataset. This can be very useful, because it allows the user to not only interpret the black box outcome for each subgroup, but also to understand the nature of the objects that a subgroup contains. We believe that this approach provides a new way to address the problem of model agnostic outcome explanation, especially when the number of outcomes to explain is large. It should be noted that although the method addresses the specific task of Incident Triage, it can be applied to any case study in the context of XAI. The whole process of the proposed approach is depicted in Figure~\ref{fig:generalSchema}.

\smallbreak
\noindent\textbf{Roadmap.} In the next section, we present our supervision and maintenance system, from rough data to black box model evaluations. Then, we introduce and formalize the novel problem of summarizing black box outcome explanations using Subgroup Discovery (Section~\ref{sec:problem}). We define the algorithm dubbed \SplitBasedSelection{} to identify subgroups along with their outcome explanations (Section~\ref{sec:solution}). We report an extensive empirical study (Section~\ref{sec:xp}) that witnesses the effectiveness of \SplitBasedSelection{} to identify interpretable subgroups with meaningful explanations in the application of incident triage. Eventually, we discuss future avenues and conclude in Section~\ref{ref:conclusion}.

\section{Methodology}~\label{sec:methodology}

\begin{table*}
    \centering
    \scalebox{1.15}{
  \begin{tabular}{c|c|c|c|c|c|c|c|c|c|c|c}
    \toprule
     \multirow{2}{*}{$O$ } & \multicolumn{2}{c|}{\textbf{r\textsubscript{title}}} & \multicolumn{3}{c|}{\textbf{r\textsubscript{summary}}} & \textbf{r\textsubscript{time}} & \multicolumn{2}{c|}{\textbf{ENV features}} & \textbf{Alerts} & \textbf{Metrics} & \textbf{Service}  \\ 
     \cmidrule(l){2-12}
     & \begin{tabular}[c]{@{}c@{}}$a_1$ \\ disk \end{tabular} & 
       \begin{tabular}[c]{@{}c@{}}$a_2$ \\ swap \end{tabular} &
       \begin{tabular}[c]{@{}c@{}}$a_3$ \\ full \end{tabular} &
       \begin{tabular}[c]{@{}c@{}}$a_4$ \\ java \end{tabular} &
       \begin{tabular}[c]{@{}c@{}}$a_5$ \\ http \end{tabular} &
       \begin{tabular}[c]{@{}c@{}}$a_6$ \\ weekend \end{tabular} &
       \begin{tabular}[c]{@{}c@{}}$a_7$ \\ Soft. version \end{tabular} &
       \begin{tabular}[c]{@{}c@{}}$a_8$ \\ Soft. type \end{tabular} &
       \begin{tabular}[c]{@{}c@{}}$a_9$ \\ Memory usage \end{tabular} &
       \begin{tabular}[c]{@{}c@{}}$a_{10}$ \\ \% used heap \end{tabular} &
       \begin{tabular}[c]{@{}c@{}}$y$ \\ class \end{tabular} \\
    \toprule 
    $o_1$     & 0.7 & 0 & 0.4 & 0 & 0 & True & 1  & Sales &  -   & 60  & \textbf{TEC}    \\ \hline
    $o_2$     & 0 & 0.8 & 0.3 & 0 & 0 & True & 3 & Sales & Blocker & 50  & \textbf{TEC}   \\ \hline
    $o_3$     & 0.5 & 0 & 0 & 0 & 0.6 & True & 2  & Factory & - & 60 & \textbf{TEC}  \\ \hline
    $o_4$     & 0 & 0.5 & 0.9 & 0.6 & 0 & True & 3  & Factory & Critical   & 97  & \textbf{OT}   \\ \hline
    $o_5$     & 0 & 0.7 & 0.6 & 0 & 0 & False & 1  & Sales & Critical  & 96 & \textbf{OT}    \\ \hline
    $o_6$     & 0.1 & 0 & 0 & 0.6 & 0.6 & False & 2  & Sales & Alarm & 85  & \textbf{OT}  \\ \hline
    $o_7$     & 0.1 & 0 & 0 & 0 & 0.9 & False & 1  & Sales & -     & 60 & \textbf{OT}    \\ 
    \bottomrule
  \end{tabular}}
  \caption{Toy Example of a dataset $(O,A,Y)$.}
  \label{fig:toyDataset}
\end{table*}

We conduct our analysis on a set $R$ of 170k incident reports that concern more than 1k servers over the last 7 years. Each server contains at least an instance of the monitored ERP software, but contains other components that are also supervised, including databases, hardware, network. We efficiently preprocess incident texts to extract discriminant features, then we can augment them with various attributes that allow to contextualize incidents, such as environment features, performance metrics, standard events, and alerts. 



\subsection{Raw data}
\noindent \textbf{Incident reports.} We define the set $R=\{r_1,...r_n\}$ of incident reports described by: (1) the title $r\textsubscript{title}$, (2) the summary $r\textsubscript{summary}$ which contains commands, stack-traces, and text written in human language (mainly French),  (3) the component $r\textsubscript{component}$ in which the incidents happen (e.g., the ERP software, the virtual machine, the storage disk), (4) the incident creation timestamp $r\textsubscript{time}$, (5) a Boolean $r\textsubscript{internal}$ that indicates whether the incident has been reported internally by our company or externally by a customer. In the latter case, we include (6) the customer $r\textsubscript{customer}$ concerned by the incident.


\smallbreak
\noindent \textbf{Environment features.} These features describe the environment and the component in which the incident happens. For instance, an ERP software is characterized by its application identifier, its version, its type among 6 main families (sales, factory, finance, etc.). A database can be described by its version (dbVersion), its maximum usage memory (sgaMax), the minimum and the maximum size of pool (jdbcMin, jdbcMax). A virtual machine can be characterized by the number of CPUs, the size of the RAM, the size of swap memory, etc.  

\smallbreak
\noindent \textbf{Performance metrics.} We continuously collect time-series that describe the behavior of several components going from hardware to business level. The usage of various memory parts is traced at a high frequency. Examples of these memory types are the heap and the non-heap for the Java Virtual Machines that run the ERP, the RAM and the swap usage, etc. Many other metrics of this kind are collected to have the maximum observability, such as the storage usage, the number of run SQL queries and their average execution time, the number of users connected to each ERP software. These metrics are of high benefit as they can efficiently indicate the context in which an incident happens, which would help for a more effective triage.

\smallbreak
\noindent \textbf{Standard events.} We collect various types of timestamped events that occur in supervised components. In the ERP software, events can result from interactions with users, such as an opening of a screen, a visualization or a creation of some data, events can also be related to automatic executions of scheduled tasks. Other components are described by different events as well: garbage collections in the Java Virtual Machine, network latency, etc.

\smallbreak
\noindent \textbf{Alerts.} Our monitoring system triggers rule-based alerts when anomalies are observed. They are generally related to a specific component and they give high indication about the location of the root cause. For example, an alert is triggered when a database tablespace is predicted to be full in less than a month. But also, another alert type identifies database table whose size increases unexpectedly. In this example, these alerts not only help to detect the problem of reaching the tablespace limit, but also locate the possible root cause, i.e., the table whose evolution leads to the tablespace saturation. Incident reports are augmented with co-occurring alert. Each alert has four levels: \textit{Info}, \textit{Alarm}, \textit{Critical} and \textit{blocker}.

\subsection{Text transformation} 
Incident titles and summaries are processed using conventional NLP techniques to extract relevant textual features. Most of these steps are achieved with the spaCy\footnote{\url{https://spacy.io/}} Python library. We start by the tokenization to extract bag of words. It is noteworthy that N-grams representation have not been used as they did not improve prediction accuracy in our use-case. We perform stemming and lemmatization. We remove special characters, stop words and noisy terms, e.g., those containing less than 3 characters, and words that occur in most of incident reports as they are not discriminant. Numbers are kept as they can be useful, such as error and response codes. Finally, we compute \textit{Term Frequency / Inverse Document Frequency} \texttt{tf-idf} of remaining terms in each incident report, and we keep only the top 10k terms in each of $r\textsubscript{title}$ and $r\textsubscript{summary}$. The number of considered terms have been chosen empirically as the one maximizing the prediction performance.

\begin{table*}
    \centering
    \scalebox{1.15}{
  \begin{tabular}{c|c|c|c|c|c|c|c|c|c|c|c|c|c|c|c}
    \toprule
     \multirow{2}{*}{ } & \multicolumn{3}{c|}{\textbf{DNN}} & \multicolumn{3}{c|}{\textbf{RF}} & \multicolumn{3}{c|}{\textbf{XGBoost}} &  \multicolumn{3}{c|}{\textbf{Naive Bayes}} & \multicolumn{3}{c}{\textbf{Logistic Regression}}  \\ 
     \cmidrule(l){2-16}
     & P & 
       R &
       F\textsubscript{1} &
       P & 
       R &
       F\textsubscript{1} &
       P & 
       R &
       F\textsubscript{1} &
       P & 
       R & 
       F\textsubscript{1} &
       P & 
       R &
       F\textsubscript{1}  \\
       \toprule 
    TOP1     & \textbf{0.84} & \textbf{0.77} & \textbf{0.82} & 0.74 & 0.72 & 0.73 & 0.75 & 0.75 & 0.74 & 0.74 & 0.75 & 0.74 & 0.75 & 0.72 & 0.73     \\ \hline
    TOP2     & \textbf{0.92} & \textbf{0.91} & \textbf{0.91} & 0.89 & 0.88 & 0.88 & 0.89 & 0.89 & 0.89 & 0.89 & 0.90 & 0.89 & 0.89 & 0.88 & 0.89    \\ \hline
    TOP3     & \textbf{0.97} & \textbf{0.96} & \textbf{0.96} & 0.94 & 0.94 & 0.93 & 0.94 & 0.94 & 0.94 & 0.95 & 0.94 & 0.94 & 0.94 & 0.94 & 0.94  \\ 
    \bottomrule
   
  \end{tabular}}
  \caption{Performance comparison of prediction models in the incident triage task.}
  \label{fig:incidentTriageScores}
\end{table*}

\subsection{A unified data structure}
We unify the different data sources into a dataset defined by a tuple $(O,A,Y)$ , where $O=\{o_i\}_{1\leq i \leq n}$ is a set of objects that refer to the historical incident reports, $A=(a_j)_{1 \leq j \leq m}$ is a vector of descriptive attributes, and the classes $Y=\{y_1,...,y_n\}$ that represent the services assigned to each incident. Each attribute $a$ can be seen as a mapping $a : O \longrightarrow R_a$ where $R_a$ is called the domain of the attribute $a$. More precisely, $R_a$ is given by $\mathbb{R}$ if $a$ is \emph{numerical}, by a finite set of categories $C_i$ if $a$ is \emph{categorical} or by $\{0, 1\}$ if $a$ is \emph{Boolean}. An incident is assigned to a service (a class) $y \in \{Class_1,...,Class_p\}$ where $p$ is the total number of services.
These notations are illustrated in Table~\ref{fig:toyDataset} with a dataset containing 7 objects $O=\{o_1,...,o_7\}$, referring to 7 incidents, each of them described by 10 attributes and a class $y$. The attributes $\{a_1,...,a_5\}$ are numerical and they provide the \texttt{tf-idf} of terms that characterize $r$\textsubscript{title} and $r$\textsubscript{summary}. The attribute ``weekend'' ($a_6$) indicates whether the incident has happened during a weekend. Other attributes correspond to some environment features, alerts, and metrics. 

\subsection{Black box model for incident triage}

The incident triage task consists in predicting the class $y$ (the service) of an incident report $r$ based on its attributes $\{a_1,...,a_m\}$. It is noteworthy that, since our goal is to make the prediction at the beginning of the incident life-cycle, all the considered attributes are available from the very first moment when the incident report is created. We consider training and validation subsets of objects $O_T, O_V \subseteq O$ that we use to train prediction models denoted by $b$. The output $b(o)=\{b(o)_1,...,b(o)_p\} \in [0,1]^p$ provides a probability distribution over the $p$ predicted output classes. Inspired by recent work~\cite{DBLP:conf/kdd/PhamJDOJ20,DBLP:conf/sigcomm/GaoYMFLBAWLMYA20,DBLP:conf/kbse/0003HL0HGXDZ19} which show in most cases that best results are achieved by DNN and ensemble models, we have evaluated the following methods: Random Forest, XGBoost, and multiple architectures of DNN. Additionally, we have considered the logistic regression and multinomial Naive Bayes as white box models to confirm that the usage of black box models effectively improve the prediction quality. We have tested each model in three different scenarios: (1) \textbf{TOP1}: predict the true service, (2) \textbf{TOP2}: predict the two most probable services, i.e., if the true service belongs to this TOP2, the prediction is considered correct, (3) \textbf{TOP3}: predict the three most probable services. Table~\ref{fig:incidentTriageScores} reports the results obtained on the validation set $O_V$. We use the most common performance metrics in multi-label classification i.e., Precision, Recall and F1 score in their weighted average formula. Interestingly, we observe that in all considered scenarios, the DNN achieves the best performances compared to other models and improves the F1 score of Logistic Regression by 9\%. In addition, the DNN is very accurate when it comes to predicting the most likely services to deal with the incident. Out of 30 different services, the model manages to reach an F-score of 0.96 when the correct service belongs to the 3 most probable services predicted by the black box model (the TOP3 setting). Therefore, we adopt the DNN model as our black box for incident triage.


\subsection{Explaining Black box outcomes}

\noindent\textbf{The single outcome explanation problem.} This problem consists in giving an explanation $e$ for the decision $b(o)=\{b(o)_1,...,b(o)_p\} \in [0,1]^p$ related to a specific object $o \in O$ where $e$ belongs to a human-interpretable domain $E$~\cite{DBLP:journals/csur/GuidottiMRTGP19}.

\smallbreak
\noindent\textbf{White box model.} In order to extract an explanation $e \in E$ for a decision $b(o)$, one of the most popular methods is to train an interpretable model that learns to imitate the decisions of $b$ specifically in the neighborhood of the object $o$. This interpretable model is called a \textit{white box model} and it is denoted $w$. New objects are synthetically generated from the neighborhood of $o$, and the model $w$ is trained to mimic decisions of $b$ on these objects. One may use linear regression as a white box model. However, since we deal with hundreds of attributes which are strongly linearly correlated, linear regression models generally tend to overfit and the model coefficients will have large variance, thus making the model unreliable. Therefore, we need to consider regularization techniques that shrink the linear model coefficients, and take into consideration the case where the number of objects to explain is less than the number of attributes used in explanation. In this paper, we use Ridge regression which penalizes the sum of squared coefficients ($L_2$ penalty). While Lasso regression is more appropriate to achieve sparsity, it has been observed that if predictors are highly correlated, the prediction performance of the lasso is dominated by ridge regression \cite{zou2005regularization}. Moreover, Lasso solution is not uniquely determined when the number of attributes is greater than the number of objects. 


\smallbreak
\noindent \textbf{Synthetic neighborhood.} For a given object $o \in O$, we denote by $N(o)$ the set of syntheticaly generated instances in the neighborhood of $o$, plus the object $o$.  To explain $o$, we train a model $w$ that imitates $b$ on the set $N(o)$. The quality of $w$ to mimic the behavior of $b$ is assessed with some fidelity measures. The fidelity can be assessed in terms of functions such as MSE and F1-score, evaluated using the outcome of the black box model  considered as an oracle. 

\begin{table*}
    \centering
    \scalebox{1.15}{
  \begin{tabular}{c|c|c|c|c|c}
    \hline
     $O$ &
     \begin{tabular}[c]{@{}c@{}}\textbf{Pred.} $b(o)_1$ \\ \textbf{of TEC} \end{tabular}&
     \begin{tabular}[c]{@{}c@{}}\textbf{Pred.} $b(o)_2$ \\ \textbf{of OT} \end{tabular}
     & \begin{tabular}[c]{@{}c@{}}\textbf{Local model} $w(o)$ \\ \textbf{of majority class} \end{tabular} & \textbf{Subgroup model} & \textbf{Subgroup description}  \\ 
       \hline 
    $o_1$     & \textbf{0.9} & 0.1 & $\text{disk}+0.1 \cdot \text{swap}+0.5 \cdot \text{full}$ & \multirow{3}{*}{\begin{tabular}[c]{@{}c@{}}$0.8 \cdot \text{disk}+0.7 \cdot \text{swap}$ \\ $+0.4 \cdot \text{full}$ \end{tabular}} & \multirow{3}{*}{$\text{weekend}=\text{True} \wedge \text{java}=0$ }     \\  \cline{1-4}
    $o_2$     & \textbf{0.8} & 0.2 & $0.1 \cdot \text{disk} + \text{swap}+0.4 \cdot \text{full}$ &  &      \\ \cline{1-4} 
    $o_3$     & \textbf{0.6} & 0.4 & $ \text{disk} - 0.5 \cdot \text{http}+0.3 \cdot \text{full}$ &  &      \\  \hline
    $o_4$     & 0.3 & \textbf{0.7} & $ 0.9 \cdot  \text{java}+0.5 \cdot \text{full} -0.2 \cdot \text{swap} $ & \multirow{2}{*}{\begin{tabular}[c]{@{}c@{}}$ \text{java}+ 0.6 \cdot \text{full} $ \\ $-0.3 \cdot \text{swap}$ \end{tabular}} & \multirow{2}{*}{$\text{\%used heap} \geq 96$ }     \\  \cline{1-4}
    $o_5$     & 0.2 & \textbf{0.8} & $  \text{java}+0.6 \cdot \text{full} - 0.3 \cdot \text{swap}$ &  &      \\  \hline
    $o_6$     & 0.2 & \textbf{0.8} & $  0.8 \cdot \text{http}$ & \multirow{2}{*}{$ 0.9 \cdot \text{http}$ } & \multirow{2}{*}{\begin{tabular}[c]{@{}c@{}}$\text{Soft. type}=\text{Sales } \wedge$ \\ $\text{weekend}=\text{False}$ \end{tabular}}     \\  \cline{1-4}
    $o_7$     & 0.1 & \textbf{0.9} & $  0.9 \cdot \text{http} - 0.1 \cdot \text{disk}$ &  &      \\  
    
    \hline
   
  \end{tabular}}
  \caption{Summarizing explanations of black box predictions of incidents from Fig~\ref{fig:toyDataset} with Subgroup Discovery. The number of explanations is reduced from 7 to 3 by grouping objects in contextualized subgroups supporting the same explanations.}
  \label{fig:toySummary}
\end{table*}
In Table~\ref{fig:toySummary}, we show $b(o)_1$ (resp. $b(o)_2$) the probability of the class TEC (resp., OT) predicted by the black box model. The majority class is TEC in $\{o_1,o_2,o_3\}$, whereas it is OT in the remaining objects. The table gives the local model $w$ trained to mimic the prediction of the majority class by the black box in the neighborhood of each object $o \in O$. For example, the local model that provides an explanation for the prediction of TEC for $o_3$ is $w(o)=\text{disk} - 0.5 \cdot \text{http}+0.3 \cdot \text{full}$. This means that the higher the \texttt{tf-idf} of the words ``disk'' and ``full'' in the incident report, the higher the probability of the TEC class, in contrast, the higher the \texttt{tf-idf} of ``http'', the lower the probability of TEC.

In practice, we may have a large set $O_E \subseteq O$ of objects whose decisions $\{b(o) \mid o \in O_E\}$ need to be explained. Providing a specific explanation for each prediction is overwhelming for the user, and it may be even impossible for her to dig into each explanation separately. In addition, many objects that share certain properties in common may have similar explanations. i.e., an explanation can be valid for a group of objects. We aim to reduce the number of explanations by partitioning the objects into subgroups $s \subseteq O_E$ that can support the same explanation. However, we need to be able to characterize these subgroups by some common interpretable description or pattern that separates them from the rest of the dataset. To this aim, we use concepts from Subgroup Discovery which are described hereafter.

\section{Summarizing black box explanations with Subgroup Discovery}~\label{sec:problem}

Subgroup Discovery (SD)~\cite{DBLP:books/mit/fayyadPSU96/Klosgen96,DBLP:journals/widm/Atzmueller15} is a data mining task that aims at identifying interpretable local subgroups that foster some properties of interest. 
This task has proven its efficiency in different applications and domains such as physics~\cite{goldsmith2017uncovering}, education~\cite{DBLP:journals/ijdsa/HelalLLEDM19} and neuro-science \cite{DBLP:journals/ploscb/LiconBSMFBGRPKB19}.
However, none of existing approaches has exploited this framework for the problem of explaining black box predictions. To make such method efficient, we needed to address several complex challenges related to the mined data structure, the subgroup interestingness, as well as a scalable mining algorithm  which optimizes an interestingness measure that is new in SD. In what follows, we start by defining subgroups and their descriptions induced by the pattern language. Then, we formally introduce our problem of summarizing black box explanations with Subgroup Discovery. 

\smallbreak
\noindent\textbf{Pattern language.} A \emph{pattern} $d$ is a \emph{constrained selector} of a subset of objects of the dataset using their attribute values. We refer to the set of all possible patterns by the \emph{pattern language} and we denote it $\mathcal{D}$. In our case, $\mathcal{D} = \bigtimes_{i=1}^m \mathcal{D}_i$ where $\mathcal{D}_i$ is given by the set of all possible intervals in $\mathbb{R}$
if $a_i$ is numerical, the set $\{C_i,\emptyset\} \cup \{\{c\} \mid c \in C_i\}$ if $a_i$ is categorical, or $\{\{0,1\},\{0\} ,\{1\}\}$ if $a_i$ is Boolean. A pattern $d \in \mathcal{D}$ is then given by a set of restrictions over each attribute (i.e. $d = (d_i)_{1 \leq i \leq m}$).  
These patterns are ordered from the most general one to the most restrictive one by an order relation $\sqsubseteq$. For two patterns $c = (c_i)_{1 \leq i \leq m} \in \mathcal{L}$ and $d = (d_i)_{1 \leq i \leq m} \in \mathcal{L}$,  we have: $c \sqsubseteq d  \Leftrightarrow \forall i \in \llbracket 1, m \rrbracket \, (c_i \supseteq d_i)$ \cite{DBLP:conf/ijcai/KaytoueKN11}.  

\smallbreak
\noindent\textbf{Linking patterns and objects.} A pattern $d = (d_i)_{1 \leq i \leq m} $ is said to \emph{cover} an object $o \in O_E$ iff $\forall i \in \llbracket 1, m \rrbracket : a_i(o) \in d_i$. A pattern $d$ \emph{covers} a set $O' \subseteq O_E$ iff it covers each object $o \in O'$. Using the \emph{cover}, we define the function $\delta(O') \in \mathcal{L}$ which gives the most restrictive pattern that covers a set of objects $O'$: $\forall d \in \mathcal{L}$, $d$ \emph{covers} $O'$ iff $d \sqsubseteq \delta(O')$. For a given pattern $d$, the set of all objects covered by $d$ is refered by $ext(d)=\{o \in O_E \mid d \sqsubseteq \delta(o)\} $ \cite{DBLP:conf/iccs/GanterK01}. In Table~\ref{fig:toyDataset}, let us simplify by considering only a subset of 4 attributes $A'=($java, weekend, Soft. type, \%used heap$)$. An example of pattern in $A'$ is $d'=(\mathbb{R}^{+},True,\text{Sales},[0,100])$, which corresponds to objects having the \texttt{tf-idf} of ``java'' in $\mathbb{R}^{+}$, ``weekend~=~True'', ``Soft.~version~=~Sales'', and values of ``\%used heap'' in $[0,100]$. This pattern $d'$ covers only $O'=\{o_1,o_2\}$. The most restrictive pattern for $O'$ in $A'$ is $\delta(O')=(0,True,\text{Sales},[50,60])$, and we have $d' \sqsubseteq\delta(O')$.

\smallbreak
\noindent\textbf{Subgroup.} A subgroup is a subset of objects $s \subseteq O_E$ that can be selected using restrictions $d$ of attributes $A$, and we note $\mathcal{S}=ext(\mathcal{L})=\{ext(d) \mid d \in \mathcal{L}\}$. In other terms, a subgroup is always characterized with some restrictions of attributes, a pattern, which makes it interpretable to a user.

\smallbreak
Instead of providing an explanation for the prediction of each $o \in O_E$, we aim to group these objects into a limited number of subgroups that cover all the objects of $O_E$ to explain, and for each subgroup, we provide an explanation that holds for all its objects. 
In Table~\ref{fig:toySummary}, we give the predicted probability $b(o)_i$ for each class. The prediction of each object $o$ is then explained by a local model $w$ trained to mimic the behavior of the black box model in the neighborhood of $o$. This model $w$ estimates the outcome of $b$ using a linear equation between \texttt{tf-idf} of terms appearing in the corresponding incident reports. 
We can partition the data into three subgroups whose objects can support the same explanation. For example, the first subgroup refers to all incidents that have happened in the weekend and that do not contain the word java in their text. Their predicted probability of TEC can be explained by the same relation: ``$0.8 \cdot \text{disk}+0.7 \cdot \text{swap}+0.4 \cdot \text{full}$''. Doing this, we summarize 7 different local models in only 3 subgroups models along with a pattern that uniquely identifies the objects explained by each model. To ensure that a subgroup model holds for all the objects of the subgroup, we seek to minimize the error made by the subgroup model while imitating the black box model on the neighborhood of each object of the subgroup. These notions are formalized below.

\noindent \textbf{Subgroup model.} A subgroup model $w_s$ is a white box model used to explain the predictions of $b$ on the objects of a subgroup $s$. It is trained on the neighborhoods of the objects of $s$. The neighborhood generation process is described later in Section~\ref{subsec:genNeigh}. 

\noindent \textbf{Loss function.}
We use the Sum of Squared Errors to evaluate the fidelity of a white box model $w_s$, fitted on a subgroup $s$ and its objects neighborhood, to imitate a black box model $b$:
$$
L(s,w_s,b)=\sum_{o \in s} \sum_{o' \in N(o)} \sum_{i = 1}^p \left( b(o')_i-w_s(o')_i\right)^2.
$$
The global loss for a set of subgroups
$S=\{s_1,s_2,...\} \subseteq \mathcal{S}$ along with their fitted models $W=\{w_{s_1},w_{s_2},...\}$ is defined as:
$
L(S,b)=\sum_{i =1}^{|S|} L(s_i,w_{s_i},b).
$

\noindent\textbf{Controlling the number of subgroups.}
To control the total number of collective explanations of the predictions $\{b(o) \mid o \in O_E\}$, we propose to upper bound the number of returned subgroups
with a threshold $K \in \mathbb{N}$. The goal is thus to find a subgroup set $\{s_1,s_2,...\}$ of size at most $K$, whose fitted white models $\{w_{s_1},w_{s_2},...\}$ minimize the loss function with respect to the black box model. The problem is formalized as follows:
\smallbreak
\noindent
\fbox{%
    \parbox{.48\textwidth}{
\begin{problem}[Summarizing explanations with SD]\label{pb:sum}
Let $O_E \subseteq O$ be a subset of objects whose predictions need to be explained, and $b$ the black box model used for prediction. Given a user-specified threshold $K \in \mathbb{N}$ representing the maximum number of explanations, find a subgroup set $S =\{s_1,s_2,... \}$ with their fitted white box models $W=\{w_{s_1},w_{s_2},...\}$ such that (1)
$|S|\leq K,$ (2)
the subgroup set covers all the objects to explain: $\bigcup_{s \in S}s=O_E,$ 
and (3) the global loss for the subgroup set is minimized: $S=argmin_{S^\prime \subseteq \mathcal{S}} L(S^\prime,b).$
\end{problem}
}}

\section{\SplitBasedSelection{} method}~\label{sec:solution}

The problem of summarizing explanations with SD is NP-Hard. This can be proven by reducing the NP-Complete problem of weighted set cover in a polynomial time to Problem~\ref{pb:sum}: each set corresponds to a subgroup, and the set weight is represented by the loss $L(s,w_s,b)$ of the corresponding subgroup. Thus, providing a scalable approach that finds the best solution to Problem~\ref{pb:sum} is not possible. We propose to use an efficient heuristic strategy detailed in Algorithm~\ref{algo:splitBasedSelection} (\SplitBasedSelection{}) and empirically prove its performance. This algorithm starts by generating the neighborhoods $N(o)$ used to train local models for each object $o \in O_E$, using \generateNeighbors{} explained in Section~\ref{subsec:genNeigh}. Then, it constructs a non-overlapping set of subgroups using a split based strategy. It begins with the subgroup set $S=\{O_E\}$ that contains a subgroup covering all the objects of $O_E$.
In each iteration, and given the current set of subgroups $S$, one of the subgroups of $S$ is split into two subgroups that minimizes the overall loss. The split is applied for one of the attributes $a \in A$. This procedure is done iteratively until the number of subgroups $K$ is reached, or, until there is no additional possible improvement of the loss, as detailed in Section~\ref{subsec:split}. 

\begin{algorithm}
    \caption{\SplitBasedSelection{}~ \label{algo:splitBasedSelection}}
  \KwIn{$O_E$ a set of objects, $b$ a black box prediction model, $K$ a threshold on the number of subgroups.}
  \KwOut{$S\subseteq \mathcal{S}$ a subgroup set that covers all the objects $O_E$, $W$ the set of white box models associated with the found subgroups.}
  {
  \For{$o \in O_E$}{
     $N(o) \gets \generateNeighbors(o)$\\
  }
  $S \gets \{O_E\}$ \\
  $W \gets dict(\{ \})$ \quad // $W$ is a dictionary \\
  $W[O_E] \gets w_{O_E}$ \quad // $ w_{O_E}$ is the white box fitted to the subgroup $O_E$ \\
  $improve \gets True$, $splits \gets \emptyset$, $newSubgroups \gets \{ O_E \}$ \\
 \While{$|S| \leq K$ and $improve$}{  \label{line:whileloop} 
// Compute the best splits for the new subgroups: \\
    \For{ $s \in newSubgroups$}{\label{line:bestSplitForS1}
    	$(a,v) \gets argmin_{a\in A, v \in R_a} L(s[a\leq v],w_{s[a\leq v]},b)+L(s[a> v],w_{s[a> v]},b)$\\
    	$splits \gets splits \cup \{(s,a,v)\}$\label{line:bestSplitForS2}\\
    	}
// Choose the subgroup split that leads to the minimum loss: \\    	
    	$(s,a,v) \gets argmin_{(s,a,v) \in splits} L(S \setminus \{s \} \cup \{s[a\leq v], s[a>v] \},b)$ \\
    	\If{$L(s[a\leq v],w_{s[a\leq v]},b)+L(s[a> v],w_{s[a>v]},b)<L(s,w_s,b)$}{
			$S \gets S \setminus \{ s \} \cup \{s[a\leq v],s[a> v] \}$   \\ 	
			remove $W[s]$ \\			
			$W[s[a\leq v]] \gets w_{s[a\leq v]}$ \\
			$W[s[a> v]] \gets w_{s[a> v]}$ \\	
			$newSubgroups \gets \{ s[a\leq v],s[a> v] \}$ \\		
			$splits \gets splits \setminus \{ (s,a,v) \}$ \\
    	}
    	\Else{
    		$improve \gets False$
    	}
 }
 return $(S,W)$\\
}
\end{algorithm}

\subsection{Neighborhood generation}\label{subsec:genNeigh}
The goal of this step is to sample a set of neighbors $N(o)$ for each object $o \in O_E$, using a locality-aware sampling strategy. Many approaches have been proposed to address this problem~\cite{DBLP:conf/kdd/Ribeiro0G16,DBLP:conf/pakdd/GuidottiMC19,DBLP:journals/corr/abs-2002-03746}. As this part of the process is not the main concern of our study, any of these approaches can be directly used in \generateNeighbors{}. However, in order to limit the bias due to this step, we use a simple yet efficient sampling approach such that (1) the closer a point $o'$ is to $o$, the higher the chance to sample it in $N(o)$, (2) the correlation between the different attributes is taken into account in order to sample more realistic objects. We first convert categorical features into numerical values. In ordinal data (e.g., Memory usage alert), while encoding, we should retain the information regarding the order in which the category is provided. Nominal features (e.g., Soft. type) are encoded so that each category is mapped with a binary variable containing either 0 or 1 using one hot encoding. In order to generate an object $o' \in N(o)$, the attribute values $A(o)$ are drawn from a multivariate normal distribution $\mathcal{N}(A,\frac{\Sigma}{z})$ centered in $A(o)$ with a covariance $\frac{\Sigma}{z}$, where $\Sigma$ is the covariance matrix of $(O_E,A)$ and $z \in \mathbb{N}$ is a parameter that shrinks the original covariance $\Sigma$ to the locality of $o$. Since the multivariate Gaussian distribution generates values in $\mathbb{R}$ for all attributes, these values need to be discretized when they correspond to non numerical attributes. Particularly, for nominal attributes, the category having the closer value to 1 among other categories of the same nominal attribute is set to 1, otherwise 0.

\subsection{Optimizing $L$ with a split-based strategy}\label{subsec:split}
Let us now detail the approach used to identify a subgroup set $S=\{s_1,s_2,...\}$ that optimizes the loss, while satisfying the constraints of coverage ($\cup_{s \in S} s=O$) and maximum size ($|S|\leq K$). In what follows, for a given subgroup $s \in \mathcal{S}$, we use the notation $s[a_i\leq v]$ and $s[a_i>v]$ to split $s$ into two subgroups with respect to the values of attribute $a_i$. By considering that $\triangleleft$ corresponds either to $\leq$ or $>$, we define $s[a_i \triangleleft v]=\{ o \in s \mid a_i(o) \triangleleft v \}$. Notice that if we split a subgroup $s$ with a Boolean attribute $a \in A$, there is only one possible split, that is $s[a_i \leq 0]$ and $s[a_i>0]$. Nominal attributes are transformed into a one hot representation, and are then treated exactly as Boolean attributes.

Algorithm~\ref{algo:splitBasedSelection} (\SplitBasedSelection{}) describes the different steps of this approach. The subgroup set is stored in $S$, and the corresponding white box models are kept in a dictionary $W$. $S$ is initialized with a subgroup $O_E$ that covers all the objects to explain. The variable $splits$ stores the best split for each subgroup $s \in S$. This variable is updated in each iteration by computing the best splits of the newly added subgroups kept in $newSubgroups$ (Line~\ref{line:bestSplitForS1} to Line~\ref{line:bestSplitForS2}). Then, the subgroup $s$ whose split reduces the loss the most is selected. It is removed from $S$ and replaced by the subgroups resulted from this split, i.e. $s[a\leq v]$ and $s[a>v]$. This loop is repeated until $|S|=K$, or until there is no further split that reduces the loss. Particularly, since the used loss function is the SSE whose optimization is convex for a linear model, a new split will either reduce $L(S)$ or let it unchanged, but it will never increase it. In fact, this is guaranteed for any model whose optimization is global, such as models with a convex loss function (linear regression, ridge regression, LASSO, etc.), as proven by the following property.

\begin{property} 
Let $s_0,s_1,s_2 \in \mathcal{S}$ s.t. $s_0=s_1 \cup s_2$ and $s_1 \cap s_2=\emptyset$, then we have: $L(s_1,w_{s_1},b)+L(s_2,w_{s_2},b) \leq L(s_0,w_{s_0},b)$.
\end{property}
\begin{proof}
This can be proven by contradiction. Let us consider that the inequality does not hold. Then, $L(s_1,w_{s_1},b)+L(s_2,w_{s_2},b) > L(s_0,w_{s_0},b)$. As $L(s_0,w_{s_0},b)=L(s_1,w_{s_0},b)+L(s_2,w_{s_0},b)$, we have  $L(s_1,w_{s_1},b)+L(s_2,w_{s_2},b) > L(s_1,w_{s_0},b)+L(s_2,w_{s_0},b)$. Two cases are then possible: 
\begin{itemize}
\item $L(s_1,w_{s_1},b)>L(s_1,w_{s_0},b)$, which means that $w_{s_1}$ is not the best model that fits $s_1$, which is absurd because $w_{s_1}$ is a global optimal solution of $L$ on $s_1$.
\item $L(s_1,w_{s_1},b) \leq L(s_1,w_{s_0},b)$, thus we have necessarily $L(s_2,w_{s_2},b)>L(s_2,w_{s_0},b)$. Following the same logic, this implies that $w_{s_2}$ is not the best fit for $s_2$, which is also absurd.
\end{itemize}
\end{proof}
\section{Experiments}~\label{sec:xp}

We report our experimental study to evaluate the effectiveness of \SplitBasedSelection{}\footnote{Source code is available on: https://github.com/RemilYoucef/split-sd4x} and its ability to summarize explanations of black box decisions in the context of incident triage. These experiments aim to answer the following questions:
\begin{itemize}
    \item \textbf{Q1}: Do subgroup models provide good explanations, in other words, are explanations of subgroup models \textit{faithful} to the black box model predictions? 
    \item \textbf{Q2}: Are subgroup models \textit{human interpretable} and do they help practitioners understand the black box results?
    \item \textbf{Q3}: Are subgroup models \textit{different} from each other?
\end{itemize}

\smallbreak
\noindent \textbf{Experiment Setup.}
We have collected 170k incident reports involving more than 1k servers over the last 7 years. Although most of data types introduced in Sec.~\ref{sec:methodology} have been used in these experiments, metrics and alerts have been omitted as their collection has been started recently and they cover only incidents of last few months. Once the data is processed and encoded, we split it randomly into training (65\%), validation (10\%) and test set (25\%). The results of the accurate black box model used for triaging are provided in Fig~\ref{fig:incidentTriageScores}. The distribution of incident reports across different services is extremely imbalanced with the most popular services having thousands of incidents while other minority services were rarely called upon. We randomly select from the test set 2000 incidents to be summarized in no more than 200 subgroups with their explanations. For that, we apply \SplitBasedSelection{} with a neighborhood size of $250$ for each object ($|N(o)|= 250$). The complete process requires about 3 hours to execute when the number of subgroups $K = 200$. Throughout these experiments, we compare \SplitBasedSelection{} with two baselines:

\begin{enumerate}
\item \textbf{Global white box} (\texttt{global-wb}): This method consists in training a white box model on the set of data that we need to explain to globally approximate the decisions of the black box model. The aim of this comparison is to see if a global white box model can effectively approximate a black box model and how much we can improve its performance with \SplitBasedSelection{}.

\item \textbf{Local white box} (\texttt{local-wb}): In this approach, we train a local white box model to explain each data object independently. We follow the same methodology applied for LIME \cite{DBLP:conf/kdd/Ribeiro0G16}, nevertheless, we opt for our proposed local neighborhood generation method to fairly compare between \texttt{SplitSD4X} and the \texttt{local-wb} model. We end up with as many models as there are objects to explain. This comparison aims to evaluate if it is possible, with a very small number of explanations to obtain similar explanatory quality results as with a large number of explanations. 
\end{enumerate}

\smallbreak
\noindent \textbf{Experiment Results.} In what follows, we present the results obtained from the evaluation of the scenarios related to the criteria defined previously.

\begin{figure}
\centering
\subfloat[\texttt{MSE} of \texttt{global-wb}, \texttt{local-wb}, and \texttt{SplitSD4X} (with different $K$)]{
\includegraphics[width=0.235\textwidth]{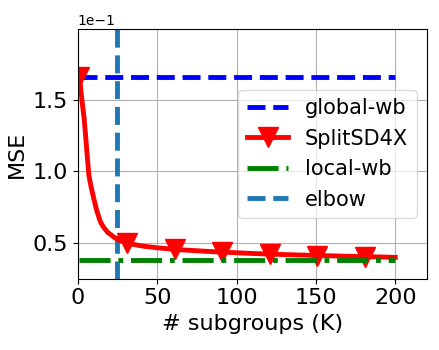}
\label{fig1:subfig1}}
\subfloat[\texttt{F1-score} of \texttt{global-wb}, \texttt{local-wb} and \texttt{SplitSD4X} with $K^{\star}$ found by \texttt{elbow} technique.]{
\includegraphics[width=0.235\textwidth]{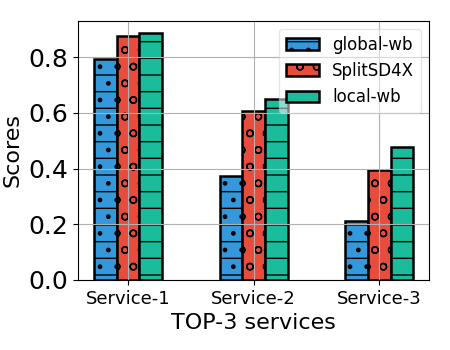}
\label{fig1:subfig2}}
\caption{\label{fig:XPfidelity}Quality of explanations of black box outcomes.}
\end{figure}

\noindent \textbf{Q1}: We evaluate whether \SplitBasedSelection{} identifies subgroups whose associated models imitate well the black box decisions. In other words, we validate whether the provided subgroup models explain the behavior of the black box while maintaining its performance. For a first analysis, we use the Mean Squared Error $\texttt{MSE}=\frac{\texttt{SSE}}{|O_E|}$ between the predictions made by \SplitBasedSelection{} and the black box model $b$ with respect to the number of computed subgroups $K$.  
Results are given in Fig.~\ref{fig1:subfig1} where we also report two constant values: the \texttt{MSE} obtained by \texttt{global-wb} and \texttt{local-wb} methods. We notice that as the number of subgroups increases, the \texttt{MSE} becomes abruptly smaller. Interestingly, the largest gain is achieved with only a small number of subgroups. To find the optimal number of subgroups $K^{\star}$, so that the fidelity is very close to that obtained by the \texttt{local-wb}, and further increasing  $K$ does not significantly improve fidelity, we use the elbow technique implemented on the available Kneed package\footnote{https://github.com/arvkevi/kneed}. We show that with only $25$ subgroup models, we can greatly improve the fidelity of \texttt{global-wb} and achieve a score quite close to that of \texttt{local-wb} which uses $2000$ models instead.    

In a second analysis, we compare \SplitBasedSelection{} with the two baselines based on the \texttt{F1-score}. These scores measure to which extent each approach is able to imitate the service predictions of the black box model. We evaluate the \texttt{F1-score} on the most 3 probable services obtained by each model compared to the results of the black box. For instance, for \textbf{Service-2}, we take the services having a second best probability and we compare them with the second probable services predicted by the black box model. Our solution is evaluated on only $25$ subgroup models. The results are shown in Fig.~\ref{fig1:subfig2}. The scores achieved by \SplitBasedSelection{} are always significantly better than the ones of \texttt{global-wb}. Furthermore, with only $25$ subgroups, we get almost similar scores results as \texttt{local-wb} (0.87 for \SplitBasedSelection{} with $25$ models and 0.88 for \texttt{local-wb} for \textbf{Service-1}). 

These results demonstrate that \SplitBasedSelection{} is able to significantly reduce the number of explanations while keeping them faithful to the black box decisions.

\begin{figure*}[htb]
\centering
\begin{center}  
 \begin{tabular}{c|c}
 
\toprule
\textbf{Sales} &  \textbf{TEC} \\
 \includegraphics[width=0.385\linewidth]{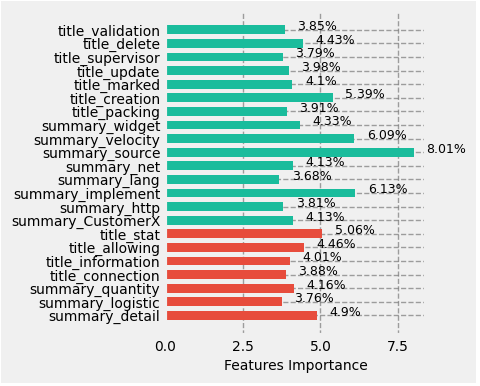} &
  \includegraphics[width=0.385\linewidth]{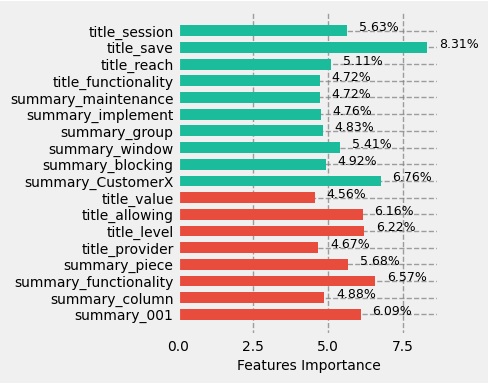}\\
 \begin{tabular}{c}
   {\scriptsize \textbf{$s_1$ description:}} {\scriptsize  $hr \in [12.0,23.0] \wedge summary\_stats = 0 $ } \\ {\scriptsize $ \wedge summary\_stock = 0$}
 \end{tabular}  & 
\begin{tabular}{c}
   {\scriptsize \textbf{$s_2$ description:}} {\scriptsize  $hr \in [1.0, 12.0] \wedge summary\_stats = 0 $ } \\ {\scriptsize $ \wedge summary\_stock = 0 \wedge app\_SV = 0$ }
 \end{tabular}
   \\

 \midrule 
 \textbf{ST} & \textbf{EDI} \\
  \includegraphics[width=0.385\linewidth]{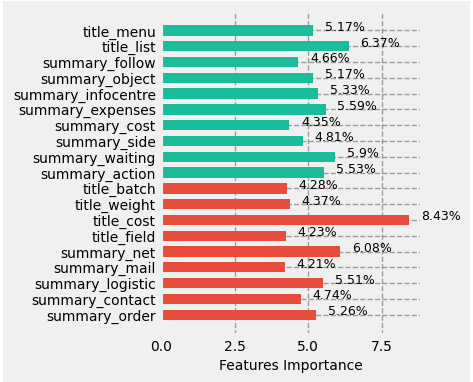}  &
 \includegraphics[width=0.385\linewidth]{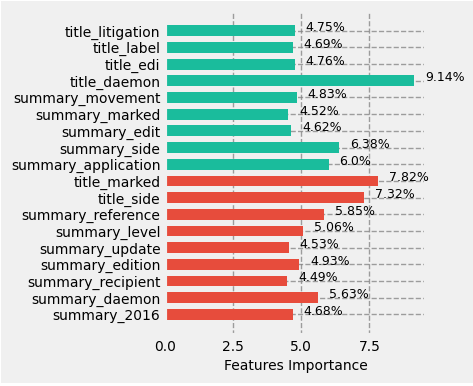}  \\
 \begin{tabular}{c}
  {\scriptsize \textbf{$s_3$ description:}} {\scriptsize  $summary\_stats = 0 \wedge summary\_stock > 0$ }
 \end{tabular}  & 

 \begin{tabular}{c}
  {\scriptsize \textbf{$s_4$ description:}} {\scriptsize  $hr \in [1.0, 12.0] \wedge summary\_stats = 0 $ } \\ {\scriptsize $ \wedge summary\_stock \wedge app\_SV = 1 \wedge day = Wednesday$ }
 \end{tabular}  \\
 \bottomrule
   \end{tabular}
\end{center}
\caption{\label{fig:sgDisplay} Subgroup examples: Patterns and the most important features of the subgroup models for specific services. Green color (resp. red) corresponds to features that contribute positively (resp. negatively) to predicting the analyzed service.}
\end{figure*}

\noindent \textbf{Q2}: The goal of our approach is to group the predicted objects into subgroups such that (1) each subgroup has an explicit description that separates it from the rest of the data, (2) the objects of the same subgroup support the same explanation. Thus, for each subgroup, we provide its description (pattern) which characterizes it, as well as its corresponding model. From this model, we derive human interpretable explanations that help practitioners to understand the reason behind predicting one service over another, by identifying the most important features of the model based on the ridge model coefficients. The contribution of each feature in predicting the analyzed class is calculated as a ratio between the absolute value of the feature coefficient over the sum of the absolute values of all model coefficients. Clearly, the relevance of feature importance depends on the fidelity of subgroup models to the black box. The better the subgroup model mimics the black box, the more we trust the coefficient-based explanation. Fig.~\ref{fig:sgDisplay} displays both the descriptions and explanations of four different subgroups, each for the prediction the most predicted service in the subgroup. We choose the most popular and requested services. For instance, the subgroup ($s_3 : summary\_stats = 0 \wedge summary\_stock > 0$) which contains reports that are characterized by a \texttt{tf-idf} of the term \textit{stock} greater than 0 and whose descriptions do not include the term \textit{stats}, is mainly dominated by the service ST that refers to Stock ($\sim 41\%$ of the incident reports). In the first subgroup, we are interested in explaining the predictions on the sales service for incidents declared between $12$ p.m. and $11$ p.m but not containing stats and stock terms in their summary. The feature importance plot highlights terms that are highly and positively correlated with the sales context such as \textit{creation}, \textit{update}, \textit{validation} and \textit{packing} of orders. While terms like \textit{velocity} which increase the probability of the sales service being requested, other terms such as \textit{logistic} and \textit{connection} decrease this probability in favor of other services. Similarly for $s_2$, we notice that each time the Technical team (TEC) has been delegated to resolve an incident, terms such as \textit{save}, \textit{session}, and \textit{blocking} are discriminating. The subgroup model of $s_3$ includes also terms that are related to stock (e.g., \textit{expenses}, \textit{cost} and \textit{menu}). The last example provided is very interesting in terms of subgroup description and quality of the associated explanation when predicting the EDI (Electronic data interchange) service. Specifically for supervision servers, we confirm that many issues related to the \textit{daemon} have been reported to this service since all EDI operations are done by daemons. Lastly, as an explanation helps us to understand and interpret the results of the black box model, it can also highlight its problems. In the second example, we notice that the term \textit{functionality} has a positive contribution in the incident title and a negative one in the summary. Such explanation helps us to better understand the behavior of the black box and try to improve it.    

\begin{figure}[ht]
\centering
\begin{tabular}{cc}
\includegraphics[width=0.32\textwidth]{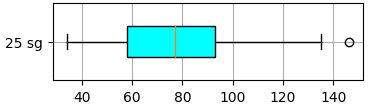}
\end{tabular}
\caption{\label{fig:distribution} Distribution of incidents on the subgroups.}
\end{figure}

\noindent \textbf{Q3}: Another question that comes to mind is whether the subgroups identified by \SplitBasedSelection{} have explanations that are different (diversified). Indeed, if most of subgroups have similar explanations, it may be still possible to summarize further these explanations, and \SplitBasedSelection{} may have failed to do it efficiently. To answer this question, we study the similarity between linear models associated to subgroups  identified by \SplitBasedSelection{} when the number of subgroups is set to $25$.
As a similarity metric between explanations of two subgroups $s_1$ and $s_2$, we use the cosine function defined between the coefficient vectors of the linear models $w_{s_1}$ and $w_{s_2}$. We have found that there is a large number of subgroups with a significant dissimilarity. Concretely, the percentage of subgroup pairs with $sim\leq 0.4$ is $93\%$. We also studied the distribution of the $2000$ incidents on the $25$ identified subgroups. As shown in Fig.~\ref{fig:distribution}, our proposed solution does not suffer from major outliers in the low end, i.e. subgroups of smallest size. On the other hand, the large number of outliers in the distribution indicates that there are lots of incidents which are similar and which must be explained simultaneously instead of treating them independently as in \texttt{local-wb}.

\section{Conclusion}~\label{ref:conclusion}
The growing use of complex equipments and softwares in modern industries conducts to the need of automating their maintenance. A cumbersome task that can be automated in maintenance is incident triage, i.e., assigning incidents to a suitable team. Black box predictive models, such as DNN, have achieved the best performance on this task. However, their obscurity strongly limits their adoption by OCEs. This has motivated our problem of explainable incident triage. We have analysed more than 170k incidents reported to our company over the last 7 years. Inspired by recent work, we have designed a black box model that is able to perform an accurate incident triage. Then, we have proposed a novel approach that provides concise explanations of our model outcomes. 

Existing methods of black box outcome explanations lead to a flood of explanations when thousands objects are predicted. In such case, analyzing individual explanations
becomes very time consuming for the end-user. To overcome this issue, we have introduced the novel problem of building explanation summaries. We have proposed a solution rooted in subgroup discovery, dubbed  \SplitBasedSelection, to group the objects whose black box prediction is supported by the same explanation. Our approach is model agnostic. The number of subgroups, and therefore the number of explanations, is controlled by a parameter. Some strategies can be defined to automatically find the relevant number of subgroups, such as the elbow method that has been explored in our experiments. Each subgroup is associated to a description that delimits the border of use of the local interpretable model which explains the black box decisions that fall into this subgroup.
Experiments carried out on incident reports demonstrate that  \SplitBasedSelection{} is able to provide a small set of high fidelity explanations of a black box model. 
Results with a small number of explanations are comparable to individual explanations of each object. For instance, 25 subgroups have been enough to explain 2,000 decisions without a significant fidelity loss. 

We believe that this work opens new directions for future research. The outcome explanation approach can be generalized to more complex configurations such as sequential models (e.g., LSTM) that progressively improve the triage when new data are available on an incident report. Another interesting direction is to extend our approach to compare model behaviors and uncover what a model captures compared to others in different situations.

\bibliographystyle{IEEEtran}
\bibliography{IEEEabrv,references}
\end{document}